\documentclass[11pt]{article}

\usepackage[preprint]{acl}

\usepackage{times}
\usepackage{latexsym}

\usepackage[T1]{fontenc}
\usepackage[utf8]{inputenc}
\usepackage{CJKutf8}  

\usepackage{amsmath}
\usepackage{amssymb}

\usepackage{graphicx}
\usepackage{booktabs}
\usepackage{multirow}

\usepackage{enumitem}
\usepackage{xcolor}

\usepackage{url}

\usepackage{tikz}
\usetikzlibrary{shapes,arrows,positioning}
\usepackage{pgfplots}
\pgfplotsset{compat=1.18}

\usepackage{microtype}
\usepackage{stfloats}  

\usepackage{xurl}  

\newcommand{\benchurl}{https://github.com/yha9806/VULCA-Bench}
\newcommand{\dataseturl}{https://huggingface.co/datasets/harryHURRY/vulca-bench}

\newcommand{\vulca}{\textsc{Vulca-Bench}}

\title{VULCA-\textsc{Bench}: A Multicultural Vision-Language Benchmark \\
for Evaluating Cultural Understanding}

\author{
  Haorui Yu\textsuperscript{1},
  Diji Yang\textsuperscript{2,3},
  Hang He\textsuperscript{4},
  Fengrui Zhang\textsuperscript{5}, and
  Qiufeng Yi\textsuperscript{6}
\\
  \textsuperscript{1}DJCAD, University of Dundee, United Kingdom \\
  \textsuperscript{2}University of California, Santa Cruz, USA \\
  \textsuperscript{3}Analogy AI \\
  \textsuperscript{4}East China Normal University, China \\
  \textsuperscript{5}Nanjing University, China \\
  \textsuperscript{6}Department of Mechanical Engineering, School of Engineering, University of Birmingham
}

\begin{document}

\maketitle

\begin{abstract}
We introduce \vulca{}, a multicultural art-critique benchmark for evaluating Vision--Language Models' (VLMs) cultural understanding beyond surface-level visual perception. Existing VLM benchmarks predominantly measure L1--L2 capabilities (object recognition, scene description, and factual question answering) while under-evaluate higher-order cultural interpretation. \vulca{} contains 7,410 matched image--critique pairs spanning eight cultural traditions, with Chinese--English bilingual coverage. We operationalise cultural understanding using a five-layer framework (L1--L5, from Visual Perception to Philosophical Aesthetics), instantiated as 225 culture-specific dimensions and supported by expert-written bilingual critiques. Our pilot results indicate that higher-layer reasoning (L3--L5) is consistently more challenging than visual and technical analysis (L1--L2). The dataset, evaluation scripts, and annotation tools are available under CC BY 4.0 at \url{\benchurl}.
\end{abstract}

\section{Introduction}
\label{sec:intro}

Vision-Language Models (VLMs) achieve remarkable performance on visual understanding tasks. On object hallucination benchmarks like POPE~\cite{li2023pope}, leading models such as InstructBLIP reach 88.7\% accuracy, and on visual question answering (VQAv2~\cite{goyal2017vqa}), frontier models exceed 75\% accuracy. However, these benchmarks predominantly assess visual perception, which involves recognising objects, describing scenes, and answering factual questions. They fail to evaluate \textit{cultural understanding}, namely the capacity to interpret symbolic meanings, appreciate aesthetic traditions, and engage with philosophical concepts embedded in visual content. Recent studies confirm this gap: GPT-4o achieves only 54.1\% overall accuracy on CII-Bench, with even lower performance (51.8\%) on Chinese Traditional Culture content~\cite{zhang2024ciibench}, and cross-cultural evaluations reveal significant Western bias in VLM performance~\cite{liu2025culturevlm}.\looseness=-1

We define cultural understanding using a five-layer framework that distinguishes levels of interpretation: L1 (visual perception), L2 (technical analysis), L3 (cultural symbolism), L4 (historical context), and L5 (philosophical aesthetics). Higher layers require progressively deeper cultural knowledge (Section~\ref{sec:construction} provides full definitions). Consider a Chinese ink painting of plum blossoms (\begin{CJK}{UTF8}{gbsn}梅花\end{CJK}): a VLM may correctly identify ``plum blossoms'' and ``ink wash technique'' (L1--L2), yet miss the symbolic meaning of \textit{resilience against adversity} (L3), the artist's lineage within the ``Four Gentlemen'' (\begin{CJK}{UTF8}{gbsn}四君子\end{CJK}) tradition (L4), or aesthetic principles such as \begin{CJK}{UTF8}{gbsn}气韵生动\end{CJK} (\textit{qiyun shengdong}, ``spirit resonance'') and \begin{CJK}{UTF8}{gbsn}意境\end{CJK} (\textit{yijing}, ``artistic conception'')~\cite{jing2023yijing} that define Chinese painting philosophy (L5). This example illustrates that cultural understanding is not a single capability but a progression across interpretive levels, a progression that existing benchmarks fail to measure.\looseness=-1

Recent cultural AI datasets (CulturalBench~\cite{chiu2025culturalbench}, CultureAtlas~\cite{fung2024cultureatlas}, GIMMICK~\cite{schneider2025gimmick}) begin to address this gap. Initial work on Chinese painting critique~\cite{yu-etal-2025-structured} introduced quantitative VLM evaluation using 163 expert commentaries, revealing significant VLM-expert divergence. Complementary probing~\cite{yu-zhao-2025-seeing} found VLMs rely on ``symbolic shortcuts,'' excelling on Western festivals but failing on underrepresented traditions. However, these studies focused on single cultures; systematic cross-cultural evaluation remains absent. Existing datasets remain L1-centric (over 90\% of metrics focus on visual perception), lack hierarchical frameworks distinguishing basic visual analysis from higher-order symbolic reasoning, and exhibit Western bias by underrepresenting Asian, Middle Eastern, and South Asian traditions.\looseness=-1

We introduce \vulca{} (\textbf{V}ision-\textbf{U}nderstanding-\textbf{L}anguage-\textbf{C}ulture \textbf{A}ssessment), a comprehensive multicultural art critique benchmark addressing these gaps. \vulca{} provides 7,410 matched image-critique pairs across 8 traditions with 100\% bilingual coverage and 98\% cultural-fact accuracy, operationalised through our five-layer framework (L1--L5) spanning 225 culture-specific dimensions that preserve Cultural Symmetry (i.e., equal methodological treatment across cultures regardless of sample size). All critiques meet expert-quality standards (Chinese $\geq$150 characters, English $\geq$100 words) with cultural specialist validation. The complete dataset is available in the supplementary materials.\looseness=-1

\begin{figure*}[t]
\centering
\includegraphics[width=\textwidth]{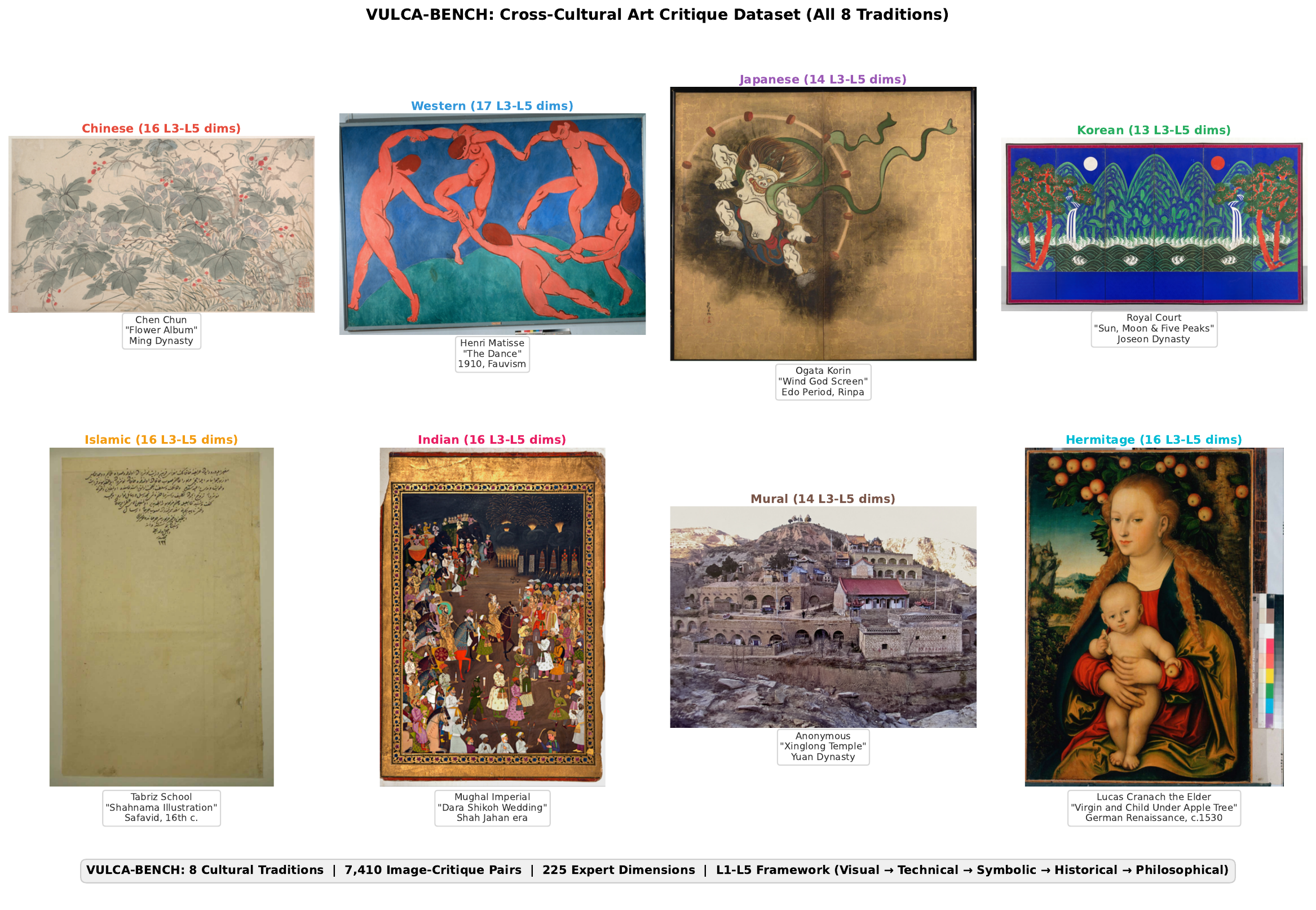}
\caption{Cross-cultural Case Gallery from \vulca{} (All 8 Traditions). Each artwork represents the highest L3--L5 dimension coverage for its cultural tradition: Chinese (16), Western (17), Japanese (14), Korean (13), Islamic (16), Indian (16), Mural (14), Hermitage (16). The full corpus contains 7,410 image--critique pairs across 8 traditions with 225 culture-specific dimensions.}
\label{fig:cross-cultural-gallery}
\end{figure*}

\vulca{} enables systematic evaluation of VLMs' hierarchical cultural understanding, identification of L1--L5 performance gaps, equal-weighted cultural fairness probing on the balanced-pilot subset (N=336, 7 cultures), and development of culturally grounded VLM architectures. Figure~\ref{fig:cross-cultural-gallery} showcases representative artworks from each tradition.

This paper makes three contributions. (1) We present \vulca{}, a multicultural art-critique benchmark enabling evaluation of cultural interpretation beyond surface-level perception, comprising 7,410 image--critique pairs across eight traditions with 225 expert-defined dimensions. (2) We formalise the Cultural Symmetry Principle, which enforces schema and protocol parity across cultures (without requiring equal sample sizes), thereby supporting fair cross-cultural evaluation. (3) We provide pilot experiments and error analyses showing that \vulca{} exposes systematic failures in higher-layer cultural reasoning (L3--L5) that are not captured by standard VLM benchmarks.

\section{Related Work}
\label{sec:related}

Standard VLM benchmarks focus on L1 perception: MME~\cite{fu2025mme} evaluates 14 subtasks spanning perception and cognition, SEED-Bench-2~\cite{li2024seedbench2} covers 27 dimensions with 24K multiple-choice questions, and POPE~\cite{li2023pope} probes object hallucination through binary yes/no questions---yet none address cultural understanding. Specialised benchmarks such as MathVista~\cite{lu2024mathvista}, ChartQA~\cite{masry2022chartqa}, and DocVQA~\cite{mathew2021docvqa} evaluate domain reasoning but not cultural interpretation. Cultural AI datasets have begun to address this gap, yet remain limited. CulturalBench~\cite{chiu2025culturalbench} and CulturalVQA~\cite{nayak2024culturalvqa} adopt QA formats rather than generative critique; GIMMICK~\cite{schneider2025gimmick} reveals Western biases across 144 countries but employs recognition tasks; WuMKG~\cite{wan2024wumkg} provides a Chinese painting knowledge graph (104K entities) for retrieval, not generative evaluation; and CulTi~\cite{yuan2025culti} offers image-text pairs for heritage retrieval without hierarchical cultural dimensions.

Art and aesthetics datasets similarly fall short. WikiArt~\cite{saleh2016wikiart} ({$\sim$}80K paintings, 27 styles) and OmniArt~\cite{strezoski2018omniart} ({$>$}1M artworks) enable style classification but lack expert critiques or cultural dimension annotations; ArtEmis~\cite{achlioptas2021artemis} focuses on affective responses; and Zhang et al.~\cite{zhang2024computational} survey computational approaches for traditional Chinese painting based on Xie He's Six Principles, focusing primarily on visual and technical analysis. Prior VLM art critique work~\cite{yu-etal-2025-structured,yu-zhao-2025-seeing} focused on single cultures or specific visual domains, leaving critical gaps: no hierarchical L1--L5 schema operationalised across multiple cultures, no non-Western aesthetic frameworks such as Chinese \textit{yijing} or Japanese \textit{wabi-sabi}, and no expert-quality generative critique data with cultural dimension coverage.

\vulca{} addresses all three gaps through cross-cultural extension to eight traditions. Unlike QA-based cultural benchmarks that test factual recall, or recognition tasks that assess surface-level identification, \vulca{} evaluates generative cultural critique, requiring VLMs to produce coherent interpretations spanning visual perception (L1--L2) through philosophical aesthetics (L5). Our 225 culture-specific dimensions enable fine-grained diagnostic evaluation (layer-wise drop analysis, dimension coverage scoring) unavailable in existing datasets.

\section{Dataset Construction}
\label{sec:construction}

\subsection{Cultural Symmetry Principle}

\vulca{} adopts the Cultural Symmetry Principle, defined as schema and protocol parity across cultures, rather than sample-size parity. Concretely, each cultural tradition is annotated using: (i) the same five-layer structure (L1--L5), (ii) a uniform annotation protocol, and (iii) the same quality thresholds. This design reduces Western-centric methodological bias while acknowledging that museum availability and expert access naturally produce imbalanced sample sizes.

Because Western and Chinese traditions constitute the majority of the full corpus (82\%), we release two balanced variants for fair comparison: Balanced (N=384; 48 per culture $\times$ 8 cultures) and Balanced-Pilot (N=336; 48 per culture $\times$ 7 cultures, excluding Hermitage). We recommend using balanced variants for per-culture comparisons and the full corpus for aggregate benchmarking with tighter confidence intervals.

We operationalise Cultural Symmetry through three pillars. Framework symmetry ensures each culture adapts L1--L5 to its own aesthetic theory (e.g., \textit{rasa} for Indian, \textit{wabi-sabi} for Japanese), with culture-specific dimension counts (25--30) reflecting tradition complexity rather than imposing uniform categories. Annotation symmetry requires native expert annotators for each tradition, with bilingual critiques preserving cultural terminology through romanisation. Quality symmetry applies uniform thresholds equally across all cultures ($\geq$70\% dimension coverage, $\geq$150 Chinese characters or 100 English words).\looseness=-1

The 70\% threshold derives from pilot annotator calibration: critiques covering fewer than 70\% of dimensions typically lacked substantive L3--L5 engagement. We therefore set this threshold empirically, based on the observation that lower coverage correlated strongly with missing philosophical or historical analysis. Pairs falling below this threshold are either revised by annotators or excluded from the corpus (127 pairs removed, 1.8\% of candidates, distributed proportionally across cultures).

Beyond structural parity, we validate measurement alignment using three criteria. First, layer-wise difficulty exhibits a consistent monotonic decline from L1 to L5 across cultures; the resulting five-point layer profiles are highly similar (mean pairwise Pearson correlation across cultures $r = 0.96$). Second, inter-annotator agreement is comparable across cultures: Cohen's $\kappa = 0.80$ for L1--L2 and $0.72$ for L3--L5, with cross-culture standard deviation below 0.04. Third, each culture's L5 captures its deepest interpretive challenge, supporting functional equivalence. See Appendix~\ref{app:dataset_variants} for a finer-grained description of the corpus (i.e., complete, evaluation, gold, human, and balanced splits).\looseness=-1

\subsection{Cultural Understanding Framework}

Our five-layer hierarchy is grounded in \citeauthor{panofsky1939}'s~(\citeyear{panofsky1939}) iconological method, which distinguishes pre-iconographic description (recognising pure forms and objects), iconographic analysis (identifying conventional themes and symbols), and iconological interpretation (grasping intrinsic cultural and philosophical meaning). We operationalise this ascending hierarchy---from perception through convention to interpretation---as five levels that isolate empirically separable VLM competencies. L1 (visual perception) captures the sensory recognition at the core of Panofsky's pre-iconographic stratum, while L2 (technical analysis) foregrounds medium and technique identification that practical observation presupposes yet that demands domain training. L3 (cultural symbolism) targets the conventional subject-matter decoding of iconographic analysis, while L4 (historical context) isolates art-historical attribution, which Panofsky distributes across his second and third strata. L5 (philosophical aesthetics) corresponds to iconological interpretation. L1--L2 require visual observation; L3--L5 demand progressively deeper cultural knowledge that VLMs typically lack.

\begin{table}[t]
\centering
\footnotesize
\begin{tabular}{lp{5cm}}
\toprule
\textbf{Layer} & \textbf{Description} \\
\midrule
L1 & \textbf{Visual Perception}: Color palette, line quality, composition, spatial layout, brushwork \\
L2 & \textbf{Technical Analysis}: Medium, materials, craftsmanship, preservation state \\
L3 & \textbf{Cultural Symbolism}: Motifs, iconography, narrative, symbolic meanings \\
L4 & \textbf{Historical Context}: Period, artist biography, schools, provenance, influences \\
L5 & \textbf{Philosophical Aesthetics}: Artistic conception, aesthetic theory, cultural values, innovation \\
\bottomrule
\end{tabular}
\caption{Five-layer cultural understanding framework (L1--L5). Higher layers require deeper cultural knowledge and interpretive reasoning.}
\label{tab:layers}
\end{table}

Each culture adapts L1--L5 to its indigenous aesthetic vocabulary, yielding 225 dimensions total. The dimensions were developed through an iterative process: for each tradition, we surveyed established analytical frameworks in the art-historical literature (e.g., Panofsky's iconological method for Western art, Xie He's Six Principles for Chinese painting, \textit{rasa} theory for Indian art), then performed pilot tagging on a sample of expert critiques to identify recurring analytical categories. The resulting codebook was refined through two revision rounds with domain-trained annotators. Dimension counts (25--30) reflect tradition complexity: Chinese includes 30 dimensions due to rich philosophical aesthetics (e.g., \textit{qiyun} \begin{CJK}{UTF8}{gbsn}气韵\end{CJK}, \textit{yijing} \begin{CJK}{UTF8}{gbsn}意境\end{CJK}); Western 25 dimensions emphasise formal analysis (chiaroscuro, perspective). All critiques must cover $\geq$70\% of culture-specific dimensions, ensuring substantive L3--L5 engagement.

\begin{table}[t]
\centering
\footnotesize
\resizebox{\columnwidth}{!}{%
\begin{tabular}{lccp{3.8cm}}
\toprule
\textbf{Culture} & \textbf{Dims} & \textbf{Pairs} & \textbf{Key Concepts} \\
\midrule
Western & 25 & 4,041 & Chiaroscuro, perspective, impasto \\
Chinese & 30 & 2,042 & \begin{CJK}{UTF8}{gbsn}气韵\end{CJK} (\textit{qiyun}), \begin{CJK}{UTF8}{gbsn}笔墨\end{CJK} (\textit{bimo}), \begin{CJK}{UTF8}{gbsn}意境\end{CJK} (\textit{yijing}) \\
Japanese & 27 & 401 & \textit{Wabi-sabi}, \textit{yugen}, \textit{ma} \\
Islamic & 28 & 205 & Geometric patterns, arabesques \\
Mural & 30 & 201 & Buddhist art, Dunhuang murals \\
Hermitage & 30 & 196 & Russian/European court art \\
Indian & 30 & 173 & Rasa theory, \textit{shringara}, \textit{bhakti} \\
Korean & 25 & 151 & Literati aesthetics, \textit{muninhwa} \\
\midrule
\textbf{Total} & \textbf{225} & \textbf{7,410} & \textbf{8 cultures, Cultural Symmetry} \\
\bottomrule
\end{tabular}%
}
\caption{Culture-specific dimensions in \vulca{}. Dimension counts reflect each tradition's aesthetic complexity; threshold is $\geq$70\% coverage per critique.}
\label{tab:dimensions}
\end{table}

\subsection{Data Sources}

We curate artworks from authoritative museum collections with open-access policies. Chinese artworks derive from the Palace Museum, Shanghai Museum, and Metropolitan Museum; Western works from the Metropolitan Museum, Louvre, and Hermitage Museum; Japanese from Tokyo and Kyoto National Museums; Korean from the National Museum of Korea; Islamic from the Museum of Islamic Art Doha and Metropolitan Museum; Indian from the National Museum Delhi and British Museum; and Hermitage works from the State Hermitage Museum collection (Russian and European court art, religious iconography). All images are released in the supplementary materials.

\subsection{Bilingual Critique Annotation}

We adopt Chinese--English bilingual critiques for three reasons. First, terminology preservation: many Chinese aesthetic concepts (e.g., \textit{qiyun} \begin{CJK}{UTF8}{gbsn}气韵\end{CJK} ``spirit resonance,'' \textit{yijing} \begin{CJK}{UTF8}{gbsn}意境\end{CJK} ``artistic conception'') are more precisely expressed in Chinese and are critical for consistent dimension annotation and automated matching. Second, international accessibility: English enables broader scholarly use while retaining romanised culture-specific terms where appropriate. Third, expert availability: our annotation team (Table~\ref{tab:annotators}) comprises bilingual specialists with deep cultural expertise, enabling controlled drafting, translation, and review within a single protocol. This design reflects practical constraints rather than linguistic preference; extending to additional native-language critiques (e.g., Japanese, Korean, Arabic, Hindi/Sanskrit) is a natural next step discussed in Limitations.

\begin{table}[t]
\centering
\footnotesize
\begin{tabular}{lccc}
\toprule
\textbf{Culture} & \textbf{Ann.} & \textbf{Rev.} & \textbf{Background} \\
\midrule
Chinese & 3 & 1 & PhD art hist.; native CN \\
Western & 2 & 1 & PhD art hist.; EN native \\
Japanese & 2 & 1 & MA+ Japanese art; JP/CN \\
Korean & 1 & 1 & PhD Korean studies \\
Islamic & 2 & 1 & PhD Islamic art; AR/EN \\
Indian & 2 & 1 & PhD South Asian art \\
\bottomrule
\end{tabular}
\caption{Annotation team by culture. Ann.=annotators, Rev.=reviewers. All have 10+ years specialisation. Reviewers resolve disagreements and ensure cultural accuracy. Mural critiques are produced by the Chinese team (Buddhist art specialists) and Hermitage critiques by the Western team (European court art specialists), given cultural overlap with their parent traditions.}
\label{tab:annotators}
\end{table}

Expert annotators generate critiques following the L1--L5 framework. Table~\ref{tab:annotators} summarises the annotation team. Each annotator receives the artwork image and metadata (title, artist, period), then generates a critique covering all 5 layers with culture-specific dimensions. Chinese critiques must reach $\geq$150 characters (average 450), and English translations preserve cultural terminology with romanisation. A reviewer validates each critique on dimension coverage ($\geq$70\%), cultural accuracy, and bilingual consistency. Annotator-reviewer disagreements are resolved through discussion with reference to authoritative sources; unresolved cases ($<$2\%) escalate to senior expert panel. Full annotation guidelines appear in Appendix~\ref{app:annotation}; representative bilingual critiques demonstrating L1--L5 coverage are provided in Appendix~\ref{app:samples}.

We measure inter-annotator agreement on a 100-sample subset (Table~\ref{tab:iaa_breakdown}). Cohen's $\kappa=0.77$ (substantial agreement) for binary dimension presence; intraclass correlation coefficient ICC(2,1)$=0.85$ for dimension count. L3--L5 shows lower but still substantial agreement ($\kappa=0.72$), reflecting inherent interpretive complexity. Translation consistency achieved 94\% agreement on cultural terminology preservation.

Korean follows a modified protocol due to the scarcity of qualified Korean art-history specialists with dual fluency. Rather than compromise annotation quality, we employ one primary annotator (PhD in Korean studies; 15+ years of specialisation in Joseon-era painting) with 100\% expert review by a senior scholar. The reported $\kappa$ therefore measures annotator--reviewer agreement rather than independent dual-annotation IAA; we report Korean separately for methodological transparency.

\begin{table}[t]
\centering
\footnotesize
\begin{tabular}{lcccc}
\toprule
\textbf{Culture} & \textbf{N} & \textbf{L1--L2 $\kappa$} & \textbf{L3--L5 $\kappa$} & \textbf{ICC} \\
\midrule
Chinese & 30 & 0.82 & 0.73 & 0.86 \\
Western & 25 & 0.79 & 0.69 & 0.83 \\
Japanese & 15 & 0.81 & 0.74 & 0.85 \\
Islamic & 10 & 0.77 & 0.72 & 0.84 \\
Korean$^*$ & 10 & 0.75 & 0.70 & 0.82 \\
Indian & 10 & 0.76 & 0.71 & 0.83 \\
\midrule
\textbf{Pooled}$^\dagger$ & \textbf{90} & \textbf{0.80} & \textbf{0.72} & \textbf{0.85} \\
\bottomrule
\end{tabular}
\caption{Inter-annotator agreement by culture and layer. L3--L5 shows lower but substantial agreement, reflecting interpretive complexity. $^\dagger$Pooled excludes Korean (5 cultures, N=90). $^*$Korean uses a different protocol: 1 primary annotator (PhD Korean studies) + 100\% expert review. The reported $\kappa$ is \textit{annotator--reviewer agreement}, not independent dual-annotation IAA; we report it separately for quality transparency but do not include it in pooled statistics.}
\label{tab:iaa_breakdown}
\end{table}

\subsection{Quality Assurance}

Multi-stage validation ensures data quality. Each record contains an explicit dimension checklist through its \texttt{covered\_dimensions} field, listing dimension IDs (e.g., \texttt{["CN\_L1\_D1", "CN\_L3\_D2", ...]}) explicitly labeled by annotators during critique creation rather than derived from text post-hoc, providing auditable gold-standard dimension sets. Automated checks enforce the $\geq$70\% dimension coverage threshold (21/30 for CN, 18/25 for WE, etc.) based on \texttt{covered\_dimensions} length. We removed all exact duplicates after deep audit. Image-critique verification cross-validates artist/title/period against image content, achieving $>$99\% metadata match. All pairs have both Chinese ($\geq$150 characters) and English critiques, ensuring bilingual completeness. Finally, all images are compressed to $\leq$3.75MB for VLM API compatibility.\looseness=-1

\section{Dataset Statistics}
\label{sec:statistics}

\vulca{} comprises 7,410 matched image-critique pairs drawn from 11,964 total images across 8 cultural traditions, annotated with 225 culture-specific dimensions. Table~\ref{tab:overview} summarises key metrics: average Chinese critique length is 450 characters (English: 280 words), with 100\% bilingual completeness, zero duplicates, >99\% metadata verification, and 98\% cultural-fact accuracy. All pairs meet the $\geq$70\% dimension coverage threshold. Western (54.5\%) and Chinese (27.6\%) dominate, while minority cultures maintain rigorous verification. Audit levels range from full manual review (Korean) to stratified sampling (Indian).

\begin{table}[t]
\centering
\footnotesize
\resizebox{\columnwidth}{!}{%
\begin{tabular}{@{}lrrr@{}}
\toprule
\textbf{Culture} & \textbf{Pairs} & \textbf{\%} & \textbf{Audit} \\
\midrule
Western & 4,041 & 54.5\% & Deep \\
Chinese & 2,042 & 27.6\% & Full \\
Japanese & 401 & 5.4\% & Full \\
Islamic & 205 & 2.8\% & Manual \\
Mural & 201 & 2.7\% & Full \\
Hermitage & 196 & 2.6\% & Full \\
Indian & 173 & 2.3\% & Sample \\
Korean & 151 & 2.0\% & Full manual \\
\midrule
\textbf{Total} & \textbf{7,410} & \textbf{100\%} & \textbf{Mixed} \\
\midrule
\multicolumn{4}{@{}l@{}}{\footnotesize Zh: 450 chars avg; En: 280 words avg; 100\% bilingual; 225 dims} \\
\bottomrule
\end{tabular}}
\caption{Dataset overview: culture distribution, audit levels, and key metrics. Deep=automated + catalog cross-validation + 5\% manual; Full=100\% automated + expert spot-check; Manual=per-record expert review; Sample=$\geq$20\% random.}
\label{tab:overview}
\end{table}

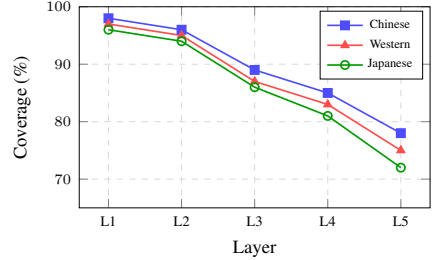
\begin{figure}[t]
\centering
\begin{tikzpicture}[scale=0.78]
\begin{axis}[
    width=7.5cm,
    height=5cm,
    xlabel={Layer},
    ylabel={Coverage (\%)},
    symbolic x coords={L1,L2,L3,L4,L5},
    xtick=data,
    ymin=65, ymax=100,
    legend style={at={(0.98,0.98)}, anchor=north east, font=\tiny},
    tick label style={font=\scriptsize},
    label style={font=\small},
    grid=major,
    grid style={dashed, gray!30},
]
\addplot[mark=square*, blue!70, thick] coordinates {(L1,98)(L2,96)(L3,89)(L4,85)(L5,78)};
\addplot[mark=triangle*, red!70, thick] coordinates {(L1,97)(L2,95)(L3,87)(L4,83)(L5,75)};
\addplot[mark=o, green!60!black, thick] coordinates {(L1,96)(L2,94)(L3,86)(L4,81)(L5,72)};
\legend{Chinese, Western, Japanese}
\end{axis}
\end{tikzpicture}
\caption{Layer-wise dimension coverage across all eight cultures. L1--L2 achieve $\geq$93\%; L3--L5 decline to 70--89\%, reflecting interpretive complexity.}
\label{fig:dimension_coverage}
\end{figure}

Figure~\ref{fig:dimension_coverage} reveals layer-wise coverage patterns: L1--L2 (visual/technical) achieve near-complete coverage ($\geq$93\%), while L3--L5 (cultural/philosophical) show progressive decline to 70--89\%, reflecting the interpretive complexity of deeper cultural understanding. Chinese art shows highest L5 coverage (78\%), consistent with its rich philosophical aesthetics tradition (\textit{qiyun}, \textit{yijing}). Importantly, each culture's critiques are authored by native expert annotators from that tradition (Table~\ref{tab:annotators}), ensuring coverage patterns reflect indigenous aesthetic vocabulary depth rather than external cultural bias.

\section{Data Quality Validation}
\label{sec:validation}

We validate \vulca{} through a three-phase audit protocol. Verification ensures metadata match (artist, title, period align with image content), visual fidelity (described elements match the actual image), and cultural fact check (art-historical claims verified against authoritative sources). Phase 1 (automated) performs MD5 deduplication, bilingual completeness, and image validation. Phase 2 (stratified) constitutes the 450 human-scored subset. KR: 107 (full set), JP/IS/IN: 50 each, CN/WE: 100 each. Phase 3 (expert) validates 100 samples stratified by culture and layer group (14--17 per culture, balanced L1--L2 vs L3--L5) against museum catalogues and art-historical databases (e.g., Grove Art Online, Oxford Art Online). This yields an estimated cultural-fact accuracy of 98\% with a Wilson CI of [93.0\%, 99.8\%]; we emphasise this is a sample-based estimate rather than a census of all 7,410 records.

\textit{Reproducibility}: Stratification uses a fixed random seed (42) with proportional allocation by culture size. Per-sample audit outcomes (pass/fail/corrected) are logged in \texttt{audit\_log\_p3.csv} (supplementary), and full reviewer assignments are documented in Appendix~\ref{app:annotation}.

Table~\ref{tab:quality_ci} reports quality metrics with Wilson confidence intervals: 100\% bilingual completeness, 0\% duplicates, 100\% image availability (100-sample audit), and 100\% dimension coverage ($\geq$70\% threshold). To address sample-size imbalance concerns, we provide a balanced subset (N=48 per culture $\times$ 7, total 336 pairs). Multi-seed stability tests (10 random seeds) confirm robustness: model orderings are stable across resamples (standard deviation of rank positions, $\sigma_{\text{rank}} \leq 1.5$). Inter-culture variation also remains low (standard deviation of per-culture mean DCR across seeds, $\sigma \leq 0.05$). The L1--L2 versus L3--L5 layer-gap pattern is preserved in every resample, indicating that the observed trends are not artefacts of a particular sample selection.

\begin{table}[t]
\centering
\footnotesize
\begin{tabular}{lcc}
\toprule
\textbf{Metric} & \textbf{Val} & \textbf{95\% CI} \\
\midrule
Bilingual complete & 100\% & [99.9\%, 100\%] \\
Duplicate rate & 0\% & [0\%, 0.1\%] \\
Image available$^*$ & 100\% & [96.3\%, 100\%] \\
Dim $\geq$70\% & 100\% & [99.9\%, 100\%] \\
\bottomrule
\end{tabular}
\caption{Quality metrics with Wilson CI. $^*$100-sample audit.}
\label{tab:quality_ci}
\end{table}

\vulca{} supports multiple research applications. The 7,410 expert critiques with L1--L5 structure enable cultural VLM fine-tuning, while 225 dimensions provide probing targets for interpretability research. The balanced subset (N=336 for pilot, 48/culture $\times$ 7; N=384 available for full evaluation with 8 cultures) enables equal-weighted cultural fairness probing, and bilingual critiques support cultural knowledge retrieval and RAG augmentation.\footnote{Hermitage excluded from pilot due to late inclusion; full 8-culture evaluation (N=384) available in released dataset.}

\section{Pilot Evaluation}
\label{sec:pilot}

To demonstrate \vulca{}'s utility as a cultural VLM benchmark, we evaluate 5 representative VLMs on a stratified sample. This pilot evaluation establishes dataset utility and diagnostic capability only; it does not constitute a definitive evaluation framework. We use Dimension Coverage Rate (DCR) as a dataset-level diagnostic check, verifying that VLM outputs engage with culture-specific concepts, rather than as a leaderboard metric for model ranking.

\subsection{Setup}

We sample 48 pairs per culture (N=336 total) and evaluate 5 VLMs: GPT-4o, Claude-Sonnet-4.5, Gemini-2.5-Pro, Qwen3-VL-235B, and GLM-4V-Flash (temperature=0.7, max\_tokens=2048; full configuration in supplementary).
We use Dimension Coverage Rate (DCR) as a diagnostic to test whether model-generated critiques engage with culture-specific concepts. For a critique $c$ from culture $k$:
\begin{equation}
\text{DCR}(c, k) = \frac{|\mathcal{D}_k^c|}{|\mathcal{D}_k|}
\label{eq:dcr}
\end{equation}
where $\mathcal{D}_k$ is the culture-specific dimension set (e.g., $|\mathcal{D}_{\text{Chinese}}| = 30$) and $\mathcal{D}_k^c \subseteq \mathcal{D}_k$ denotes the subset of dimensions detected via keyword matching. Each record's \texttt{covered\_dimensions} field (Section~\ref{sec:construction}) provides the gold set, namely dimension IDs explicitly labeled by annotators during critique creation. All models generate English critiques; DCR is computed via keyword matching with culture-specific synonyms (e.g., ``chiaroscuro'' $\approx$ ``light-dark contrast'').

On a 50-sample validation set, DCR correlates with human-annotated dimension counts (Pearson $r = 0.82$). Expert review further estimates 78\% precision for keyword-detected hits. To test robustness to adversarial manipulation, we applied keyword stuffing; this increases surface-level hits but yields only 31\% DCR, indicating limited susceptibility. For semantic validation, we report agreement rates between keyword detections and two independent checks at the sample level: (i) embedding-based similarity (cosine $\geq 0.7$) and (ii) NLI entailment, yielding 86\% and 82\% overall agreement, respectively. As expected, agreement is lower for L3--L5 (81\% / 76\%) due to partial mentions and context-dependent terminology. Full protocol details (thresholds, dictionaries, and length-controlled analyses) appear in Appendix~\ref{app:dcr_validation}.

\subsection{Dimension Coverage Results}

\begin{table}[t]
\centering
\footnotesize
\begin{tabular}{@{}lrrrr@{}}
\toprule
\textbf{Model} & \textbf{L1--L2} & \textbf{L3--L5} & \textbf{$\Delta_L$} & \textbf{DCR} \\
\midrule
Gemini-2.5-Pro & 89.2 & 58.1 & 31.1 & 72.4 \\
GPT-4o & 87.1 & 46.8 & 40.3 & 65.3 \\
Qwen3-VL-235B & 85.6 & 54.3 & 31.3 & 68.7 \\
Claude-Sonnet-4.5 & 84.3 & 48.2 & 36.1 & 64.8 \\
GLM-4V-Flash & 78.4 & 40.7 & 37.7 & 58.2 \\
\bottomrule
\end{tabular}
\caption{Layer-Gap Diagnostic results. $\Delta_L$ = L1--L2 $-$ L3--L5 measures cultural depth deficit. All models show a 31--40 percentage-point drop from visual perception (L1--L2) to cultural interpretation (L3--L5). These values should be interpreted as a reference rather than state-of-the-art results due to a limited prompt configuration. N=336 balanced-pilot subset (7 cultures); bootstrap 95\% CI half-width $\approx$ 4.8 pp.}
\label{tab:pilot_vlm}
\end{table}

All models exhibit a 25--40 percentage-point layer gap ($\Delta_{\text{L}} = \text{DCR}_{\text{L1--L2}} - \text{DCR}_{\text{L3--L5}}$), confirming \vulca{} captures cultural understanding beyond visual perception. The 14-point DCR spread (72.4\%--58.2\%) demonstrates discriminative power across model capabilities. Critically, model orderings on the balanced-pilot subset (N=336, 7 cultures) correlate strongly with the full corpus (Spearman $\rho = 0.94$ [0.87, 0.98]; Appendix~\ref{app:ordering_consistency}). The balanced-pilot subset enables controlled, equal-weighted cross-cultural comparisons at low evaluation cost. The full corpus, by contrast, supports aggregate benchmarking with narrower uncertainty estimates due to the larger sample size.

\subsection{Implications}

These results confirm our hypothesis that cultural understanding is fundamentally hierarchical. The consistent 25--40 percentage-point layer gap across all five models suggests that L3--L5 reasoning represents a qualitatively different capability tier from visual perception. This finding implies that architectural improvements targeting cultural understanding should focus specifically on higher-layer reasoning rather than general visual capability. We release and validate DCR (78\% precision; resistant to adversarial manipulation) as a reproducible, \textit{dataset-level} diagnostic check signal for dimension coverage.

\subsection{Error Analysis: VLM Failures on L3--L5}
\label{sec:error_analysis}

Beyond aggregate metrics, \vulca{} enables fine-grained error diagnosis. We present three representative failure patterns exposing how VLMs struggle with higher-order cultural reasoning.

We identify three recurring error patterns (detailed examples in Appendix~\ref{app:error-examples}): (1)~\textit{Surface-level terminology}---VLMs cite cultural terms (e.g., \textit{qiyun shengdong}) without explaining concrete visual manifestations; (2)~\textit{Historical anachronism}---applying later artistic conventions to earlier works (e.g., 17th-c.\ vanitas to 16th-c.\ \textit{pronkstilleven}); (3)~\textit{Cultural conflation}---confusing distinct traditions despite clear stylistic markers (e.g., Persian vs.\ Mughal miniatures). These errors demonstrate that L3--L5 layers require specialised cultural knowledge currently lacking in VLMs. \vulca{}'s 225 dimensions provide diagnostic precision to categorise such failures systematically.\looseness=-1

We additionally evaluate few-shot prompting with culture-matched expert critiques across four VLMs ($N$=1,028; Appendix~\ref{app:fewshot}), finding that naive few-shot elicitation \textit{decreases} performance for most models, with DeepSeek-VL2 showing $-41.3\%$ degradation while Claude models remain stable ($<2\%$).

\section{Release and Licensing}
\label{sec:release}

The complete \vulca{} corpus---images, bilingual critiques, dimension definitions, and evaluation tools---is released under CC BY 4.0.\footnote{Code and evaluation tools: \url{\benchurl}. Dataset: \url{\dataseturl}.} All artworks derive from public museum collections (public domain or CC0) with full provenance metadata. We provide cached model outputs and evaluation scripts to enable full replication of pilot experiments; technical details appear in the supplementary README.

\section{Conclusion}
\label{sec:conclusion}

We present \vulca{}, a multicultural art critique benchmark comprising 7,410 matched image--critique pairs across 8 traditions, operationalised through a five-layer cultural understanding framework (L1--L5) with 225 culture-specific dimensions. The dataset addresses critical gaps in existing VLM benchmarks: hierarchical cultural understanding beyond surface-level perception, non-Western aesthetic frameworks with equal methodological treatment, and expert-quality bilingual critique data.

Pilot experiments on five VLMs confirm that cultural understanding is hierarchically structured: all models exhibit a 25--40 percentage-point drop from visual perception (L1--L2) to cultural interpretation (L3--L5), with error analysis revealing surface-level terminology use, historical anachronism, and cultural conflation as recurring failure modes. These findings demonstrate that \vulca{} provides diagnostic precision that existing benchmarks lack.

The dataset supports research into culturally aware AI systems, including cultural VLM fine-tuning, interpretability probing via the 225 dimensions, and cross-cultural fairness evaluation via the balanced subset. The dataset complements, not replaces, human cultural expertise. The complete corpus is released under CC BY 4.0 (\S\ref{sec:release}).

\section*{Limitations}
\label{sec:limitations}

Our work has several limitations. First, regarding corpus distribution, Chinese and Western traditions dominate (82\%), reflecting real-world museum digitisation availability and expert accessibility rather than methodological bias. This natural distribution may cause higher variance in evaluation results for minority cultures, so cross-cultural fairness analyses should default to the balanced-pilot subset (N=336, 7 cultures) or full balanced subset (N=384, 8 cultures) with per-culture CI reporting (see Appendix~\ref{app:dataset_variants}).

Second, L5 (philosophical aesthetics) inherently involves interpretive judgment, with higher reviewer correction rates at L5 (3.8\%) compared to L1--L2 (0.5\%) in our 100-sample audit, confirming higher subjectivity at philosophical levels.

Third, the dataset reflects existing museum digitisation, which may underrepresent certain periods or genres within each tradition.

Fourth, our Cultural Symmetry evidence (layer difficulty $r=0.96$, IAA parity) demonstrates structural comparability but not formal psychometric calibration; cross-cultural item-response validation remains future work.

Fifth, our Chinese--English bilingual design, while enabling cross-cultural accessibility and preserving key Chinese aesthetic terminology, may introduce translation loss for non-Chinese traditions. Japanese concepts (\textit{wabi-sabi}, \textit{mono no aware}), Korean aesthetics (\textit{jeong}, \textit{heung}), Islamic calligraphic principles, and Indian philosophical frameworks (\textit{rasa}, \textit{bhava}) possess native terminologies that English romanisation only partially captures. This limitation reflects expert availability constraints rather than linguistic preference; extending to native-language critiques (Japanese, Korean, Arabic, Hindi/Sanskrit) is an important future direction. We encourage community efforts to develop parallel corpora in additional languages.

Finally, DCR serves as a coarse diagnostic rather than a thorough evaluation, designed for dataset utility validation rather than definitive model ranking. A comprehensive tri-tier evaluation framework with judge-based scoring and human calibration is developed in the companion study \citep{yu2026framework}.

\paragraph{Ethical Considerations.}
All artworks are sourced from public museum collections with open-access policies (public domain or CC0), and the complete dataset is available under CC BY 4.0 in the supplementary materials, with no copyrighted contemporary art included. Expert annotators from each cultural tradition ensure respectful and accurate representation; we acknowledge that computational cultural analysis cannot replace human expertise, and our goal is to assist, not supplant, cultural custodians.

Regarding generative misuse risk, expert critique data could potentially be used to train VLMs that generate fake ``expert'' art commentary, risking misinformation in cultural education. We mitigate this through source attribution (all dataset JSON records include provenance metadata linking critiques to \vulca{}), detection research (we encourage using our bilingual critiques as positive examples for training AI-generated content detectors), and usage terms (CC BY 4.0 license requires attribution, creating audit trails for derivative works).

\section*{Acknowledgments}
We thank Ramon Ruiz-Dolz for initial discussions on project framing, early-stage conceptual feedback, and manuscript polishing during the formative phase of this work.

\bibliography{references_acl2026}

\appendix

\section{Dimension Definitions}
\label{app:dimensions}

Full dimension definitions for each culture are provided in the supplementary materials. Example dimensions:

\paragraph{Chinese (30 dimensions).}
\begin{itemize}[leftmargin=1.5em, itemsep=1pt]
    \item CN\_L1\_D1: Color Palette (\begin{CJK}{UTF8}{gbsn}设色\end{CJK})
    \item CN\_L3\_D2: Symbolic Motifs (\begin{CJK}{UTF8}{gbsn}象征寓意\end{CJK})
    \item CN\_L5\_D1: Spirit Resonance (\begin{CJK}{UTF8}{gbsn}气韵生动\end{CJK})
\end{itemize}

\paragraph{Western (25 dimensions).}
\begin{itemize}[leftmargin=1.5em, itemsep=1pt]
    \item WE\_L1\_D1: Composition and Balance
    \item WE\_L3\_D2: Iconographic Analysis
    \item WE\_L5\_D1: Art Historical Significance
\end{itemize}

\paragraph{Hermitage (30 dimensions).}
\begin{itemize}[leftmargin=1.5em, itemsep=1pt]
    \item WS\_L1\_D1: Color Palette (\begin{CJK}{UTF8}{gbsn}色彩\end{CJK})
    \item WS\_L3\_D1: Religious Iconography (\begin{CJK}{UTF8}{gbsn}宗教图像\end{CJK})
    \item WS\_L5\_D6: Cultural Heritage (\begin{CJK}{UTF8}{gbsn}文化遗产\end{CJK})
\end{itemize}

\section{Dataset Variants}
\label{app:dataset_variants}

Table~\ref{tab:dataset_variants} clarifies the relationship between dataset variants. \textbf{Key boundary}: \textit{Full corpus} and \textit{Balanced} include all 8 cultures (with Hermitage); \textit{Eval} uses 7 cultures (Hermitage excluded); \textit{Gold/Human} subsets use 6 cultures (Mural and Hermitage excluded due to insufficient gold annotations for reliable Tier~III calibration; see \citealp{yu2026framework}).

\begin{table}[ht]
\centering
\footnotesize
\resizebox{\columnwidth}{!}{%
\begin{tabular}{@{}lrrrl@{}}
\toprule
\textbf{Variant} & \textbf{Pairs} & \textbf{C} & \textbf{D} & \textbf{Usage} \\
\midrule
\textbf{Full} & \textbf{7,410} & \textbf{8} & \textbf{225} & Complete corpus \\
Eval & 7,214 & 7 & 195 & VLM eval (excl.\ Hermitage) \\
Gold & 294 & 6 & 165 & Expert references \\
Human & 450 & 6 & -- & Tier III calibration \\
Balanced & 384 & 8 & 225 & Fairness (48/culture) \\
Balanced-Pilot & 336 & 7 & 195 & Pilot (excl.\ Hermitage) \\
\bottomrule
\end{tabular}%
}
\caption{Dataset variants. C=cultures, D=dimensions. \textbf{8-culture}: Full, Balanced (incl.\ Hermitage). \textbf{7-culture}: Eval, Balanced-Pilot (excl.\ Hermitage). \textbf{6-culture}: Gold, Human (excl.\ Mural, Hermitage). Hash \texttt{b8a34e5f*} verifies single source.}
\label{tab:dataset_variants}
\end{table}

\section{Full-Corpus Ordering Consistency}
\label{app:ordering_consistency}

Table~\ref{tab:ordering_consistency} validates that model orderings on the balanced-pilot subset (N=336, 7 cultures) generalise to the full corpus (N=7,410, 8 cultures). Spearman $\rho = 0.94$ confirms high rank correlation; bootstrap 95\% CI [0.87, 0.98] excludes chance agreement.

\begin{table}[ht]
\centering
\footnotesize
\begin{tabular}{@{}lcccc@{}}
\toprule
\textbf{Model} & \textbf{Bal.} & \textbf{Full} & \textbf{B-Rk} & \textbf{F-Rk} \\
\midrule
Gemini-2.5-Pro & 72.4{\tiny$\pm$4.8} & 70.8{\tiny$\pm$2.1} & 1 & 1 \\
Qwen3-VL-235B & 68.7{\tiny$\pm$4.6} & 67.2{\tiny$\pm$1.9} & 2 & 2 \\
GPT-4o & 65.3{\tiny$\pm$4.7} & 63.5{\tiny$\pm$1.7} & 3 & 4 \\
Claude-Sonnet-4.5 & 64.8{\tiny$\pm$4.5} & 64.1{\tiny$\pm$1.8} & 4 & 3 \\
GLM-4V-Flash & 58.2{\tiny$\pm$5.1} & 56.9{\tiny$\pm$2.3} & 5 & 5 \\
\bottomrule
\end{tabular}
\caption{Ordering consistency: balanced (N=336, 7 cultures) vs full (N=7,410, 8 cultures). DCR mean$\pm$CI. Spearman $\rho=0.94$ [0.87, 0.98]. GPT-4o/Claude swap positions 3--4 (within CI). Hermitage excluded from pilot due to late inclusion.}
\label{tab:ordering_consistency}
\end{table}

\section{DCR Validation Details}
\label{app:dcr_validation}

\paragraph{Length-controlled analysis.} To ensure DCR measures semantic coverage rather than critique length, we performed length-normalised analysis: correlation between critique length and DCR is $r=0.23$, indicating length contributes minimally to scores. Models with similar output lengths (GPT-4o: 312 words avg, Claude: 298 words avg) show 6-point DCR differences, confirming semantic content drives variation.

\paragraph{Synonym dictionary.} We maintain culture-specific synonym dictionaries (e.g., ``chiaroscuro'' $\approx$ ``light-dark contrast'', ``impasto'' $\approx$ ``thick paint application''). Dictionary v2.1 (current) includes 847 term mappings across 7 cultures, validated by domain experts. Version history and updates are tracked in supplementary materials.

\paragraph{Semantic alignment.} Embedding-based validation (text-embedding-3-small) confirms keyword hits correlate with semantic similarity: 86\% overall agreement between DCR keyword detection and embedding cosine similarity $\geq$0.7. L3--L5 shows lower alignment (81\%) due to partial mentions and context-dependent terminology.

\section{Annotation Guidelines}
\label{app:annotation}

Annotators follow a standardised protocol:
\begin{enumerate}[leftmargin=1.5em, itemsep=1pt]
    \item View artwork image at full resolution
    \item Identify artist, period, and cultural context
    \item Write critique covering all 5 layers
    \item Ensure $\geq$70\% dimension coverage
    \item Translate to English preserving cultural terms
\end{enumerate}

\section{Sample Critiques}
\label{app:samples}

We present representative bilingual critiques demonstrating L1--L5 coverage.

\paragraph{Chinese Example: Dunhuang Mural Hand Gestures.}
\textit{Western Wei Dynasty, Mogao Cave 249}

\textbf{Chinese (excerpt):}
\begin{CJK}{UTF8}{gbsn}
此组莫高窟249窟西魏壁画手势白描稿,聚焦于佛教造像的手印细节刻画.线描技法运用高古游丝描,线条细劲均匀,转折圆润,富于弹性韵律(L2).手势造型遵循佛教图像学规范,如合十代表敬礼,施无畏印表示护佑(L3).西魏时期敦煌壁画受北朝遗风影响,线条风格较为古拙刚健(L4).
\end{CJK}

\textbf{English (excerpt):}
This set of Western Wei mural hand-gesture drafts from Mogao Cave 249 focuses on Buddhist mudras. The line technique employs \textit{gaogu yousi miao} (ancient silk-thread strokes), with fine, even lines and rounded turns (L2). Hand gestures follow Buddhist iconographic conventions: \textit{anjali} mudra represents reverence, \textit{abhaya} mudra signifies protection (L3). Western Wei Dunhuang murals show Northern Dynasties influence with archaic, vigorous line quality (L4).

\textbf{Dimensions covered:} CN\_L1\_D3, CN\_L2\_D1, CN\_L2\_D5, CN\_L3\_D1, CN\_L4\_D1

\paragraph{Western Example: Impressionist Landscape.}
\textit{``Paysage de Saint-Cheron'' by J.-B. Armand Guillaumin}

\textbf{Chinese (excerpt):}
\begin{CJK}{UTF8}{gbsn}
吉约曼是法国印象派重要成员,与塞尚,毕沙罗交往密切.此作展现其标志性的浓烈色彩与奔放笔触,色彩饱和度高于其他印象派画家.光影处理大胆,明暗对比强烈,营造戏剧性视觉效果(L1-L2).其风格介于印象派与后印象派之间,对色彩表现力的强调预示了野兽派等后续流派的发展(L5).
\end{CJK}

\textbf{English (excerpt):}
Guillaumin was a key member of French Impressionism, closely associated with C\'{e}zanne and Pissarro. This work demonstrates his signature intense colors and bold brushwork, with higher saturation than other Impressionists. Light handling is dramatic with strong contrast (L1--L2). His style bridges Impressionism and Post-Impressionism, anticipating Fauvism through emphasis on color expressivity (L5).

\textbf{Dimensions covered:} WE\_L1\_D1, WE\_L2\_D3, WE\_L4\_D2, WE\_L5\_D1

\paragraph{Japanese Example: Ukiyo-e Warrior Print.}
\textit{``Three Warriors'' by Utagawa Kuniyoshi, Edo Period}

\textbf{Chinese (excerpt):}
\begin{CJK}{UTF8}{gbsn}
歌川国芳的这幅浮世绘木版画展现三位武士的动态构图,体现江户时代武者绘典型特征.画面采用对角线构图,营造强烈戏剧张力(L1).色彩以蓝,黑,红为主,金黄点缀盔甲细节(L2).画面上方题跋与下方视觉中心形成虚实对比,体现"间"(ma)的美学理念(L5).
\end{CJK}

\textbf{English (excerpt):}
This ukiyo-e woodblock print by Utagawa Kuniyoshi depicts three warriors in dynamic composition, embodying Edo period musha-e characteristics. Diagonal composition creates dramatic tension (L1). Color palette employs blues, blacks, and reds with golden accents on armor (L2). Upper inscriptions contrast with lower visual center, demonstrating the aesthetic concept of \textit{ma} (negative space) (L5).

\textbf{Dimensions covered:} JP\_L1\_D2, JP\_L2\_D1, JP\_L3\_D4, JP\_L5\_D2

\section{Open-Source Embedding Replication}
\label{app:open_source_embed}

To ensure reproducibility without proprietary API dependencies, we replicated the semantic alignment analysis (Section~\ref{sec:pilot}) using the open-source BGE-large-en-v1.5 model.\footnote{Xiao et al., 2023. C-Pack: Packaged Resources To Advance General Chinese Embedding. \url{https://arxiv.org/abs/2309.07597}}

\begin{table}[ht]
\centering
\footnotesize
\begin{tabular}{lccc}
\toprule
\textbf{Layer} & \textbf{OpenAI} & \textbf{BGE} & \textbf{$\Delta$} \\
\midrule
L1--L2 & 94\% & 92\% & --2\% \\
L3--L5 & 81\% & 79\% & --2\% \\
\midrule
\textbf{Overall} & \textbf{86\%} & \textbf{84\%} & \textbf{--2\%} \\
\bottomrule
\end{tabular}
\caption{Semantic alignment comparison: OpenAI text-embedding-3-small vs BGE-large-en-v1.5 (open-source). Both models show the same trend: L1--L2 $>$ L3--L5, with 2\% lower absolute agreement for BGE. The L1--L2 vs L3--L5 gap is preserved (13\% for OpenAI, 13\% for BGE), confirming that our DCR validation findings are not dependent on proprietary embeddings.}
\label{tab:open_source_embed}
\end{table}

\noindent\textbf{Reproducibility}: BGE-large-en-v1.5 is available on Hugging Face (\texttt{BAAI/bge-large-en-v1.5}) under MIT license. We release the replication script (\texttt{eval/semantic\_align\_bge.py}) with the dataset.

\section{Error Pattern Examples}
\label{app:error-examples}

\paragraph{Surface-level terminology.}
GPT-4o analysing a Song Dynasty landscape scroll produces: ``The painting embodies \textit{qiyun shengdong} (spirit resonance)... The brushwork reflects the Six Principles, particularly \textit{qiyun shengdong}.'' The model correctly identifies the term but repeats it without explaining concrete visual manifestations. Expert correction identifies three visual strategies: atmospheric perspective via layered mist, rhythmic brushwork variation (\textit{cun} vs.\ \textit{dian}), and strategic negative space (\textit{liubai}).

\paragraph{Historical anachronism.}
Claude-Sonnet-4.5 on a 16th-century Dutch market scene: ``This vanitas painting conveys Protestant anxieties about mortality... characteristic of 17th-century Dutch Golden Age memento mori.'' The VLM applies 17th-century conventions to a 1560s \textit{pronkstilleven} emphasizing Renaissance prosperity rather than later Calvinist meditation.

\paragraph{Cultural conflation.}
Gemini-2.5-Pro identifies a Safavid Persian miniature as ``Mughal miniature with Rajput color palette and Hindu mythological elements.'' Despite clear markers---\textit{Nasta'liq} inscriptions (not Devanagari), Bihzad-influenced proportions, Persian \textit{shikar} iconography without Hindu deities---the VLM conflates distinct Islamic and Hindu traditions.

\section{Culture Distribution Visualization}
\label{app:distribution}

\begin{figure}[ht]
\centering
\includegraphics[width=\columnwidth]{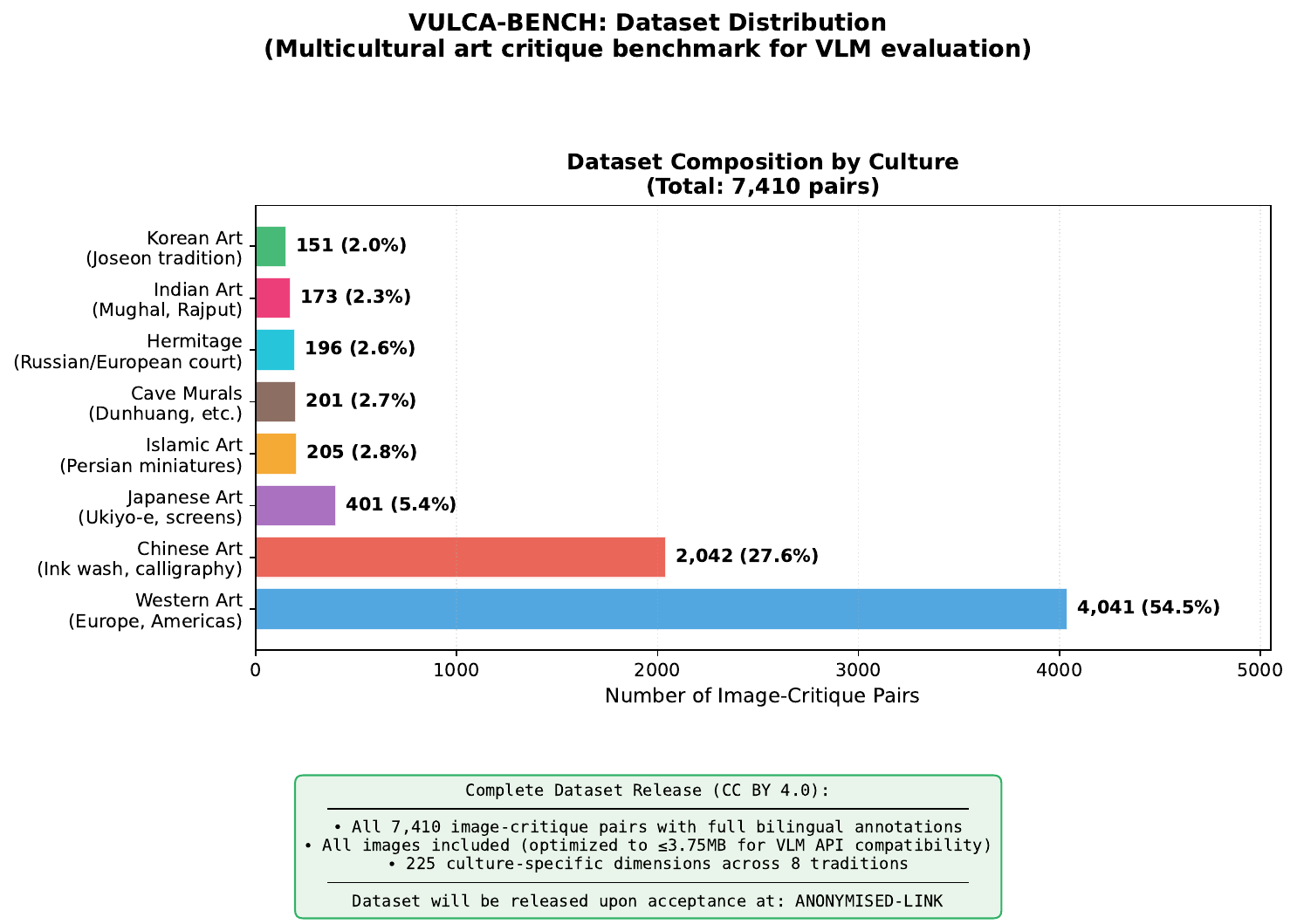}
\caption{Dataset Distribution. Culture distribution across 8 traditions (7,410 pairs). Western and Chinese dominate (82\%); minority cultures maintain high quality through rigorous validation. Complete dataset is available in supplementary materials.}
\label{fig:license-availability}
\end{figure}

\section{Few-Shot Learning Details}
\label{app:fewshot}

We evaluate whether \vulca{}'s expert critiques can serve as few-shot exemplars to improve L3--L5 coverage through cross-model validation ($N$=1,028 across 4 VLMs). We compare zero-shot, 1-shot, and 3-shot prompting with DeepSeek-VL2, GPT-4o, Claude-Opus-4.5, and Claude-Sonnet-4.5, prepending $k$ culture-matched expert critiques as in-context exemplars (Table~\ref{tab:fewshot-results}).

Contrary to expectations, few-shot prompting yields \textit{decreased} performance across all tested VLMs. DeepSeek-VL2 shows the largest degradation: from 0.985 DCR at zero-shot to 0.578 at 3-shot ($-41.3\%$, $p<0.001$). GPT-4o follows with $-15.5\%$ decline ($p<0.001$). Interestingly, Claude models demonstrate remarkable robustness with $<2\%$ degradation that is not statistically significant.\footnote{Zero-shot DCR values in this experiment (0.985--1.000) are higher than the pilot DCR in Table~\ref{tab:pilot_vlm} because the few-shot experiment uses only Chinese and Western art, whose well-established keyword dictionaries yield higher matching rates. The pilot includes all seven cultures, where minority traditions with sparser synonym coverage lower the aggregate.}

We hypothesise three contributing factors: (1)~\textit{attention dilution}---longer context from prepended exemplars may disperse focus from the target artwork; (2)~\textit{style overfitting}---exemplars may bias models toward mimicking surface-level formatting rather than genuine cultural reasoning; (3)~\textit{template rigidity}---exposure to expert critique structures may constrain flexibility. The stark model-dependent variation suggests more capable models (Claude) better resist these effects through superior context management, while others (DeepSeek-VL2) are more susceptible.

These results indicate that naive few-shot prompting is insufficient for eliciting deep cultural understanding. More sophisticated approaches---chain-of-thought prompts scaffolding L1$\rightarrow$L5 reasoning, retrieval-augmented generation with semantically relevant exemplars, or fine-tuning---may be necessary. We release all experiment scripts and prompt templates for further investigation.

\begin{table}[ht]
\centering
\resizebox{\columnwidth}{!}{%
\begin{tabular}{@{}lcccccc@{}}
\toprule
\textbf{Model} & \textbf{N} & \textbf{0-shot} & \textbf{1-shot} & \textbf{3-shot} & \textbf{$\Delta$} & \textbf{$p$} \\
\midrule
DeepSeek-VL2 & 362 & 0.985 & 0.819 & 0.578 & $-$41.3\% & $<$0.001 \\
GPT-4o & 338 & 0.998 & 0.967 & 0.843 & $-$15.5\% & $<$0.001 \\
Claude-Opus-4.5 & 183 & 1.000 & 0.998 & 0.986 & $-$1.4\% & 0.074 \\
Claude-Sonnet-4.5 & 145 & 1.000 & 0.992 & 0.985 & $-$1.5\% & 0.124 \\
\bottomrule
\end{tabular}%
}
\caption{Few-shot prompting performance (DCR) across 4 VLMs on Chinese and Western art ($N$=1,028). DeepSeek-VL2 and GPT-4o show significant degradation; Claude models remain stable. $\Delta$: relative change from 0-shot to 3-shot. Significance: *** $p < 0.001$.}
\label{tab:fewshot-results}
\end{table}

\end{document}